\pdfoutput=1
\documentclass[letterpaper, 10 pt, conference]{ieeeconf}  % Comment this line out if you need a4paper

\IEEEoverridecommandlockouts                              % This command is only needed if 
\usepackage{graphicx}
\usepackage{adjustbox}
\usepackage{hhline}
\usepackage{multicol}
\usepackage{multirow}
\usepackage{threeparttable}
\usepackage{amsmath}

\title{\LARGE \bf
Learn to Quantify Social Interaction with Constraints for Pedestrian Walking
}

\author{ Xiaodan Shi
\thanks{Xiaodan Shi is with Department of Computer and Systems Sciences, 
    Stockholm University
        {\tt\small xiaodan.shi@dsv.su.se}}%
\thanks{Corresponding Author: Xiaodan Shi}% <-this % stops a space
}

\begin{document}

\maketitle
\thispagestyle{empty}
\pagestyle{empty}

%%%%%%%%%%%%%%%%%%%%%%%%%%%%%%%%%%%%%%%%%%%%%%%%%%%%%%%%%%%%%%%%%%%%%%%%%%%%%%%%
\begin{abstract}
Long-term human path forecasting in crowds is  critical for  autonomous moving platforms (like autonomous driving cars and social robots) to avoid collision and make high-quality planning.  Although the current research take into account social interactions for prediction, they don't reveal the exact kinds of social interactions happened among people and how the social interactions affect the decision-making process of pedestrians, which further limits its robustness.  Social interactions in pedestrian walking are intuitively massive and hard to label and quantify. In this paper, we explore creatively to quantify and interpret how pedestrians interact with others by proposing \textit{Learn to Cluster}. Our clustering social interactions is probabilistic latent variable generative, learning directly from sequential trajectory observations,
scalable to arbitrary number of pedestrians. Learn to cluster is label-free and can be naturally integrated into the training process of the prediction model. The latent variables will then serve as 'labels' to categorize social interactions.  Extensive experiments over several trajectory prediction benchmarks demonstrate that our method is able to learn the patterns of social interactions and effectively integrate the patterns to pedestrian trajectory prediction.
\end{abstract}

%%%%%%%%%%%%%%%%%%%%%%%%%%%%%%%%%%%%%%%%%%%%%%%%%%%%%%%%%%%%%%%%%%%%%%%%%%%%%%%%
\section{Introduction}\label{intro}
Learning social etiquette lies at the heart of  understanding human trajectories in crowded scenes. For example, the pedestrians or bicyclists can quickly integrate all the surrounding  pedestrians, bicyclists or cars and yield the right-of-the-way. Pedestrians naturally have learnt the ways how to interact with physical space and other moving objects by avoiding collisions and respecting the social distance. Learning those principles are not only important for understanding pedestrians behaviors but also helps autonomous vehicles and robots to navigate safely in very complex scenes with extraordinary proficiency. 

Understanding how pedestrian socially interact with others in crowds is at the core of trajectory prediction problem. Trajectory prediction is a problem that predicts future positions of pedestrians or vehicles given past trajectory observations, has raised much attentions in recent years due to its potential applications on many transportation applications. The models of trajectory prediction are usually agent-based, tackling the social interactions to make more accurate predictions by integrating the hidden moving states of surrounding objects through a graph structure. 

Social interactions among crowds are complicated, heterogeneous and hard to interpret. Despite a large body of works on modeling social interactions for trajectory prediction, the existing research usually visually analyse the social behaviors in afterwards experiments as avoid collisions, walking together and merging into a group etc.  However, no exclusive  information about the types of social interactions are truly considered in networks.  Trajectory prediction models usually  blindly pool all kinds of social interactions without considering their distinction, which means the networks modeling social interactions are not intepretable. The types of social interactions and how they affect the decision-making process of pedestrians aren't revealed yet through the current research. 

\begin{figure}
    \centering
    \includegraphics[clip=true, viewport=50 240 600 600, width=0.40\textwidth]{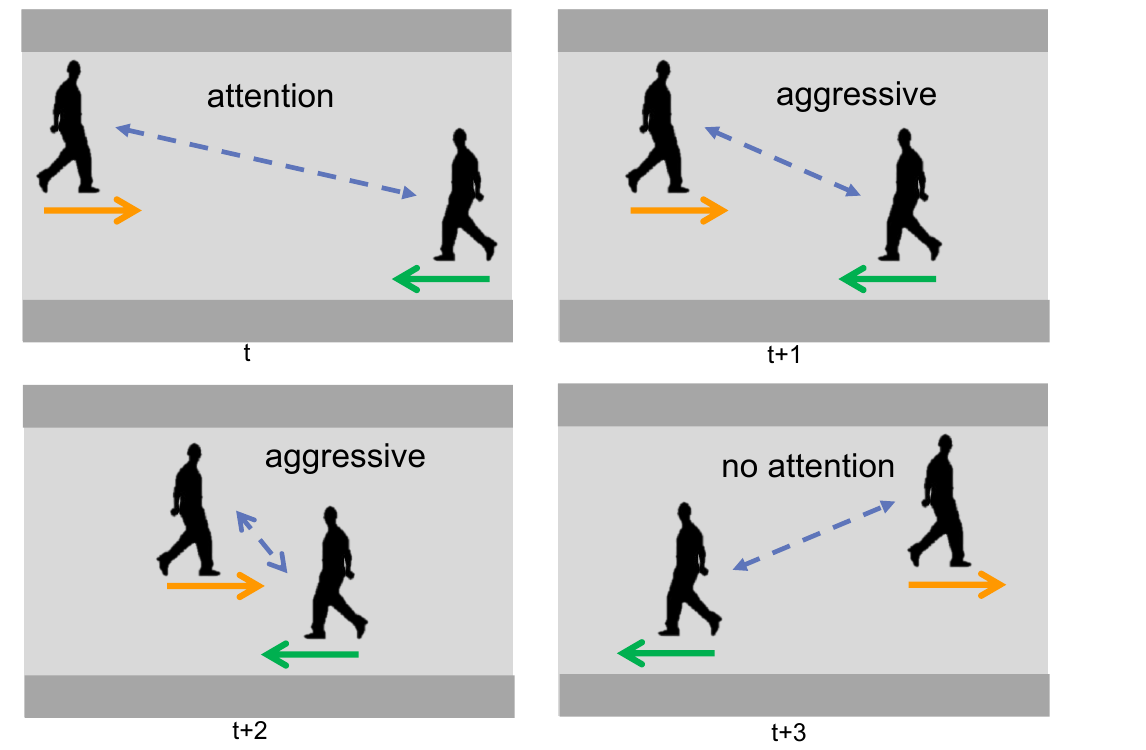}
    \caption{From a common-sense perspective, social interactions can be broadly categorized into aggressive, mild, and no interaction. The scenarios illustrated represent relatively simple cases, in which the presence or absence of social interaction can often be readily identified. However, in more complex scenarios, it becomes much more difficult to determine whether social interactions occur.}
\end{figure}

In this paper, we shed light on the interpretability of social interactions for trajectory prediction, creatively revealing how pedestrians  exactly socially interact with others and their decision-making process for navigation through crowds. To achieve this, a unifying framework is proposed to describe the possible types of social interactions. Our framework modeling social interactions is probabilistic latent variable generative, learning directly from sequential trajectory observations, scalable to arbitrary number of pedestrians.  Our model takes on the point-of-view of any agent and aggregates the relative motions between agent and its neighbor pedestrians through a common-used graph structure. Latent variables further automatically learns to describe the  possible socially meaningful modes of social interactions in a probability way without explicit labeling. In fact, our modeling social interactions clusters all the social interactions through the discrete latent variables, where social behaviors in the same cluster should be  similar and in the different cluster should be distinctive. To capture this, we  also propose a social loss function to regulate the inter- and cross-cluster  "distances" to maximize the cross-cluster similarities and to minimize the inter-cluster similarities. 

Besides, we also propose a simple yet efficient method for integrating the learned modes of social interactions to prediction model. In the prediction model, we  adopt Mixture Density Network (MDN) to generate multiple future positions and compute the loss over all components of the mixture model for capturing the uncertainty of prediction. We test the model using classic trajectory prediction benchmarks.  
To address the challenge of interpreting the learned clusters, we propose an intuitive strategy by designing several common-sense interaction scenarios to indirectly explain the meaning of these modes. The experiments show promising and comparable results.

% Use the style of numbering in square brackets.
% If nothing is used, default style will be taken.
%\begin{enumerate}[a)]
%\item 
%\item 
%\item 
%\end{enumerate}  

%\begin{itemize}
%\item 
%\item 
%\item 
%\end{itemize}  

%\begin{description}
%\item[]
%\item[] 
%\item[] 
%\end{description}  

% Figure
% \begin{figure}[<options>]
% 	\centering
% 		\includegraphics[<options>]{}
% 	  \caption{}\label{fig1}
% \end{figure}

% \begin{table}[<options>]
% \begin{tabular*}{\tblwidth}{@{}LL@{}}
% \toprule
% \midrule
%  & \\
%  & \\
%  & \\
%  & \\
% \bottomrule
% \end{tabular*}
% \end{table}

% Uncomment and use as the case may be
%\begin{theorem} 
%\end{theorem}

% Uncomment and use as the case may be
%\begin{lemma} 
%\end{lemma}


\section{Related Works}\label{related}
\subsection{social Interactions for Trajectory Prediction}
Social LSTM introducing Social Pooling to learn a global feature of all nearby neighbors around an agent which is meant to represent common sense rules and social conventions, is a tipping point for data-driven long-term trajectory prediction. Many research follow the way of Social LSTM\cite{alahi2016social} but with improvements. Attention mechanism is introduced to learn neighbors' weights on agent~\cite{sadeghian2019sophie,bae2023eigentrajectory,marchetti2024smemo}. Fernando et al. extended the classic model to incorporate
both soft attention as well hard attention where the former
is for handling longer trajectories and the latter is used for
modeling interacting people~\cite{fernando2018soft+}. Instead of directly modeling hidden states of neighbors' motion, some research pool relative motion between agent and neighbors to model interactions. SR-LSTM proposed a state refinement module for LSTM, which extracting social effects of neighbors by embedding and aggregating the relative spatial location between agent and neighbors~\cite{zhang2019sr}.
Graph representation, specifically spatio-temporal graph (ST-graph) is well applied to illustrate human motion and their interactions~\cite{shi2020multimodal,mohamed2020social,yu2020spatio,peng2021stirnet}.  ST-graph provide a more direct and natural way to model interactions for trajectory prediction. Structure-RNN~\cite{jain2016structural} combining high-level spatio-termporal graphs with sequence modeling success of RNN made significant improvements on problem of human motion modeling. Some research follow this direction. 
Social-BiGAT introduced a flexible graph attention network to model social interactions between pedestrians in a scene. It assumes all people in a scene interacting instead of setting a local neighborhood~\cite{karasev2016intent}. Social-STGCNN utilized spatio-temporal graph representation and proposed a weighted
adjacency matrix to meansure the influence between pedestrians~\cite{mohamed2020social}. Recently, Transformer is also  used to model the motion and social interactions for trajectory prediction~\cite{li2020end,yuan2021agentformer,liu2021multimodal}. Li etc. utilized self-attention mechanism to integrate social interactions by using queries Q to represent the agent actor, keys K and values V to represent neighbor agents~\cite{li2020end}. Although most of the current research claim they consider social interactions for future prediction, it is hard to say what kind of social interactions  going on among pedestrians are really encoded. Thus in the paper, we investigate to explain the social netiquettes among pedestrians and to encode  the explainable social interactions for prediction problem.

\subsection{Multi-modality of Trajectory Prediction}
Human motions under crowded scenarios imply a multiplicity of modes. To capture the uncertainty of future path, some research apply generative adversarial network (GAN) or variable autoencoder (VAE)  to generate multiple possible paths ~\cite{gupta2018social,sadeghian2019sophie,cheng2021amenet,chen2021personalized,neumeier2021variational}. Gupta A. et al. proposed Social GAN which contains RNN based encoder-decoder generator and RNN-based decoder discriminator~\cite{gupta2018social}. Social GAN integrates all the interactions involved in the scenarios and encourages the generative network to spread its distribution and cover the space of possible paths by introducing a variety loss. Sadeghian A. et al proposed Sophie, an attentive GAN to jointly model static human-space, and dynamic human-human interactions by blending a social attention mechanism with a physical attention that helps the model to learn where to look in a large scene and to extract the most salient parts of the image relevant to the path~\cite{sadeghian2019sophie}. 
Some research apply Mixture Density Network (MDN) to map the distribution of future trajectories~\cite{shi2020multimodal,bishop1994mixture,makansi2019overcoming,eiffert2020probabilistic} . The article~\cite{makansi2019overcoming}, based on MDN, proposed a two stage strategy that first predicted several samples of future with Winner-Takes-All loss and then iteratively grouped the samples to multiple modes.
There are also goal-based multi-trajectory prediction~\cite{tang2019multiple,zhang2020map,gu2021densetnt,zhao2021you,girase2021loki}. Those models predict multiple futures based on hypothesis of goals. One kind of goal-based prediction models the trajectories based on the semantic destinations, such as turning right/left, going straight~\cite{tang2019multiple,li2020evolvegraph}. Another kind firstly  forecasts multiple positional designations and then estimates futures matching the goal hypothesis~\cite{dendorfer2020goal}. We also model the multi-modality of trajectory and forecast multiple plausible futures by using MDN. But worth noting that it is not our key contribution and we mainly focus on modeling explainable social interactions.  We predict multiple futures mainly for: (1) to better compare our method with other baselines; (2) to demonstrate the proposed explainable social interactions able to apply to forecast multi-modal futures. 

\begin{figure*}
    \centering
    \includegraphics[clip=true, viewport=0 280 1150 800, width=0.95\textwidth]{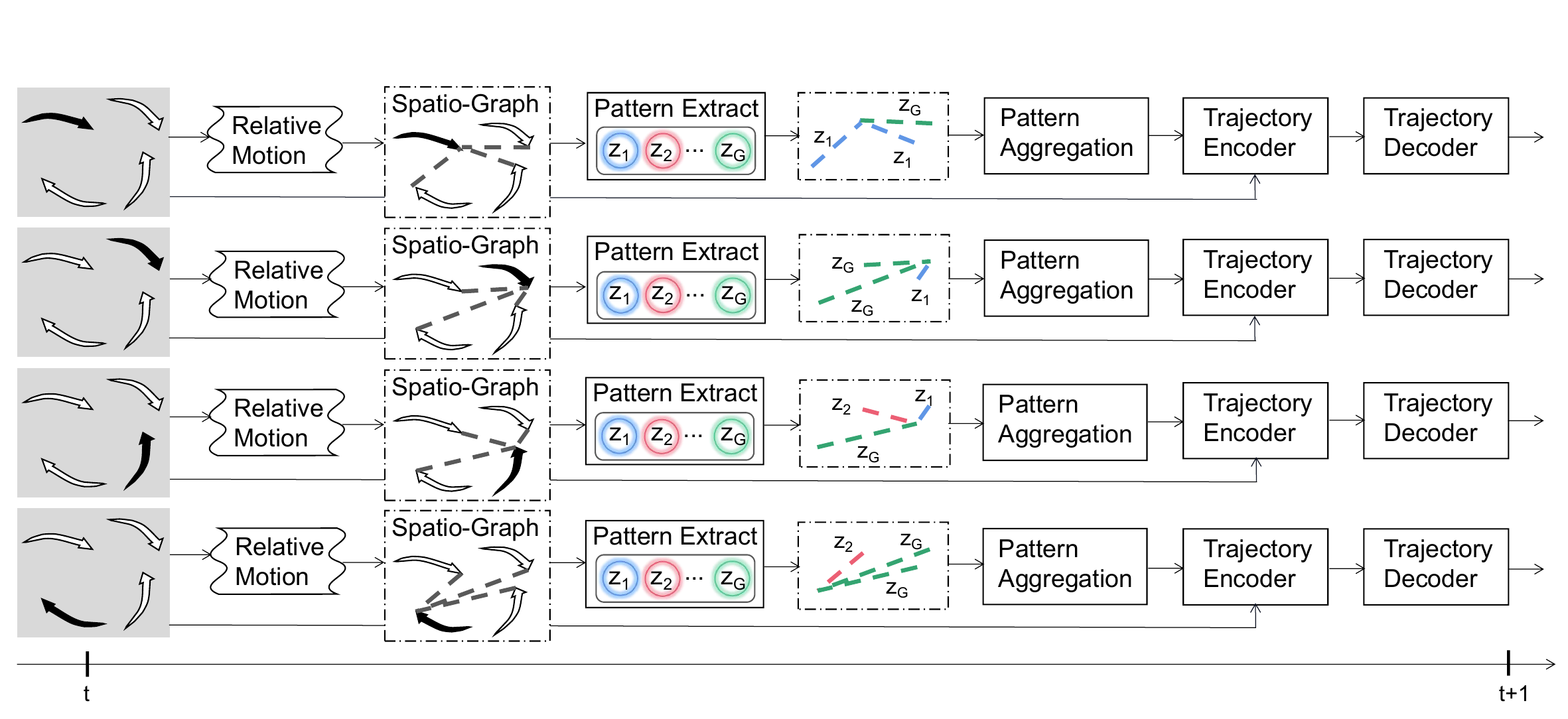}
    \caption{Model Architecture. We first employ a spatio-temporal graph~\cite{mohamed2020social} to jointly represent pedestrian interactions and walking trajectories. Within the prediction network, latent variables are learned from the features of social interactions to capture their clustered representations. The social interaction features, together with the corresponding cluster information, are then integrated into the prediction process through a pattern aggregation module.}
    \label{fig:modearchitecture}
\end{figure*}

\section{Methodology}\label{method}
\subsection{Problem Formulation}\label{problem}
We assume that each scenario has been preprocessed to get 2D spatial coordinates $(x_i^t, y_i^t)\in \textbf{R}$ and 2D walking speed $ (u_i^t, v_i^t)\in \textbf{R}$ of all people at all time instances. 
There are  $N$ agents in a scenario. The observation of agent $i$ is  past trajectories represented as: $X_i^{1:\tau} = \{(x_i^t, y_i^t, u_i^t, v_i^t)| t = 1,2, \cdots ,\tau \}$ while the future trajectories is $Y_i^{\tau:T} = \{(x_i^t, y_i^t)| t = \tau+1, \cdots , T \}$. 

Our goal is to learn the posterior distribution $p(Y_i^{\tau:T}|X_i^{1:\tau}, X_{1:N \backslash i}^{1:\tau})$. To generate the distribution of future trajectories, we jointly model multiple ego-trajectories and their interactions with $f$. Therefore, the distribution is denoted as:
\begin{equation}
    p(Y_i^{\tau:T}|X_i^{1:\tau}, X_{1:N \backslash i}^{1:\tau}) \,  = \, f(X_i^{1:\tau}, X_{1:N \backslash i}^{1:\tau};w^{*}),
\end{equation}
where $w^{*}$ are the parameters of the model we aim to learn. We denote the predicted future paths as $\hat{Y}^{\tau:T}$ whose distributions are learned from our model.

%; uniformly in British English is preferred.

%------------------------------------------------------------------------- 

\begin{figure}
    \centering
    \includegraphics[clip=true, viewport=150 230 600 650, width=0.25\textwidth]{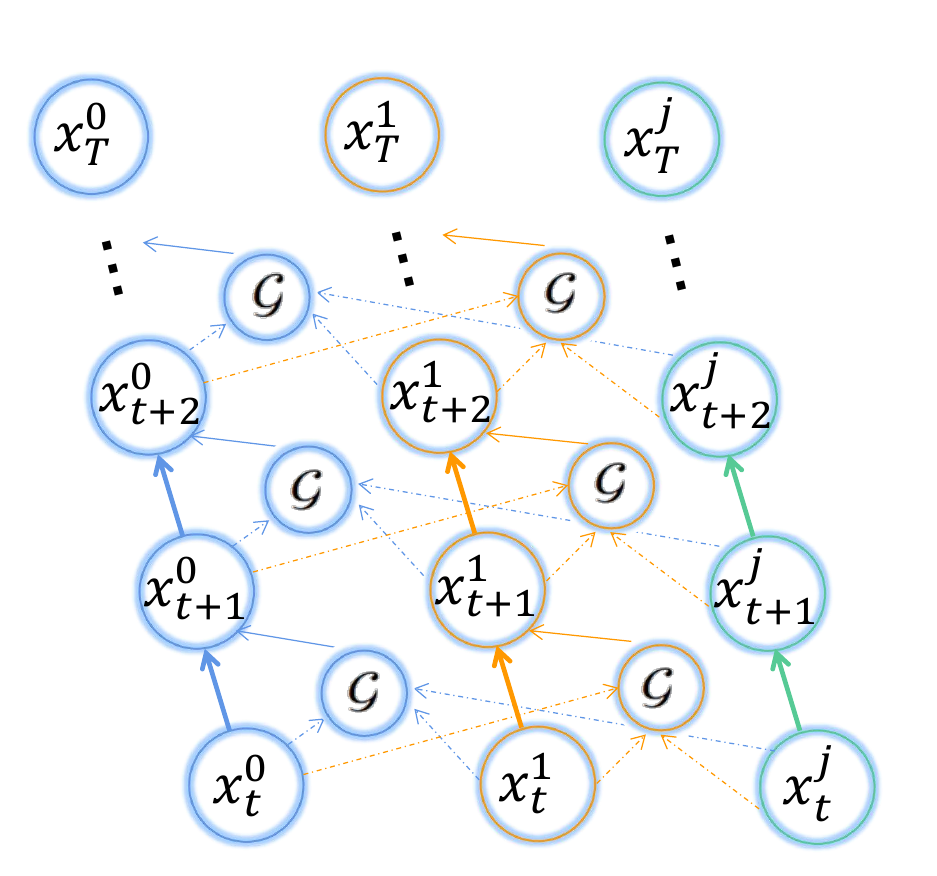}
    \caption{Calculation of mode. We calculate the mode of social interactions at every time instance.}
    \label{fig:cofmode}
\end{figure}

\subsection{Social Interactions}\label{interaction}
As shown in Figure~\ref{fig:modearchitecture}, all the pedestrians at any time instance $t$ are characterized with location $(x^t, y^t)$ and offset $(u^t, v^t)$.  Offsets are common-used states describing trajectories, which can stabilize the training  process and improve the model performance. We assume the agent pedestrian is indexed as $i$ and its social graph is constructed by formulating relative motions between it and all the neighbors in a neighborhood setting. We use $\mathcal{E}_{j}^{t}$ to denote the spatial edge between $i$ and neighbor $j$ as follows:
\begin{equation}
\begin{array}{l}
    \mathcal{E}_{i,j}^{t} = [(x_i^t, y_i^t)-(x_j^t, y_j^t),(u_i^t, v_i^t)-(u_j^t, v_j^t) ]
\end{array}
\end{equation}
We assume there are $M$ neighbors of the agent at time instance $t$. $\mathcal{E}_{i}^{t}=\{\mathcal{E}_{i,j}^{t}|j=1, 2,\cdots, M\}$, has the size of (M, 4). $M$ is dynamic, free to be any number as long as the neighbor pedestrians are in the neighborhood of the agent. 
$\mathcal{E}_{i,j}^{t}$ is then  embedded with a fully connected layer with ReLU activation and featured as $\hat{\mathcal{E}}_{i,j}^{t}$.

To interpret and extract patterns of social interactions, we introduce a set of stochastic latent variables $\mathcal{G}_\sim Multinoulli(G)$ that is directly embedded in the prediction network and conditioned on the latent features of social interactions and  will learn  to indicate the similarity and discrimination between massive dynamic interactions. For each pairwise agents at time instance $t$, $\mathcal{G}_{i,j}$ is conditioned on $\hat{\mathcal{E}}_{i,j}^{t}$ and can take on $G$ discrete values to define the social interaction patterns of the spatial edge (Figure ~\ref{fig:cofmode}). Those discrete  variables are finite and latently label each social interaction directly from the trajectory data. % The learned latent variable-represented patterns don't have exact meanings initially. 
We understand social interactions with the same latent labels lie within the same  cluster, sharing the similar social behaviours, owning the same interaction pattern. Likewise, social interactions with different variables have different social behaviours.  The social interactions patterns/clusters described by  variable $\mathcal{G}$ don't match the explicit meanings initially. The patterns don't mean the exact social behaviours such as merging into a group, walking parallel etc., but tend to indicate the styles of social behaviours such as aggressive: agents pay attention to those neighbors, try to avoid collision with them and adjust path mainly based on their interactions,  mild:  agents pay attentions to  those neighbors, keep safe distance with them and adjust path partly based on their interactions,  no attention: pay no attention to  those neighbors, etc. As we mentioned before, we further construct a tensor 
 $\hat{\mathcal{E}}_{i}^{t}$  by embedding each element of $\mathcal{E}_{i}^{t}$ to more succinctly get the embedding features of our latent patterns by using a trick similar to how Transformer model get multi-head attentions. $\hat{\mathcal{E}}_{i}^{t}$ has original size of (M, R*G) where $R*G$ is the embedding size of pairwise relative motion $\hat{\mathcal{E}}_{i,j}^{t}$, is then reshaped as (M, G, R).  We use a softmax function to $\hat{\mathcal{E}}_{i}^{t}$ to get discrete probability distribution of social interaction patterns as follows:
\begin{figure}
    \centering
    \includegraphics[clip=true, viewport=150 380 600 650, width=0.40\textwidth]{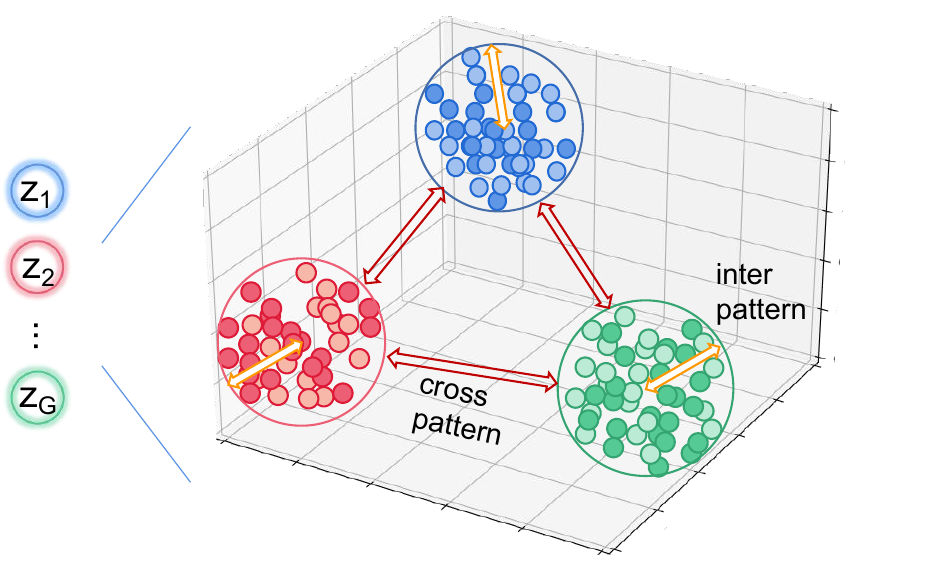}
    \caption{Regulations on social patterns. The features of same clusters contract and features of different clusters contrast.}
\end{figure}

%  We utilize latent parameters $\mathcal{G}\sim Multinoulli(G)$ to denote interaction patterns between two agents (types of $\hat{\mathcal{E}}^{t}$). For each pairwise agents at time instant $t$, $\mathcal{G}$ is conditioned on $\hat{\mathcal{E}}^{t}$ and can take on $G$ discrete values. Further, we describe interaction patterns as a discrete probability distribution given by:

\begin{equation}
\begin{array}{l}
    p(\mathcal{G}_i^t|X_{1:M}^{t})=Softmax(\hat{\mathcal{E}}_i^{t})
\end{array}
\end{equation}
where, $p(\mathcal{G}_i^t|X_{1:M}^{t})$ is the probability distribution over $G$ interaction patterns and reflects the possibilities of patterns each social interaction belong to.  To get the exact social pattern, we need to sample from  the discrete probabilities. However, the sampling process in the middle of the network lead to the problem of backpropagation interruption due to its non-differentiability. To address it, we further utilize Gumbel-Max, the reparameterization trick  to sample from $p(\mathcal{G}_i^t|X_{1:M}^{t})$.
% \begin{equation}
% \begin{array}{l}
%     \mathcal{G}_j^t = onehot(Softmax(log(p(\mathcal{G}_j^t|X_{1:N}^{0:t}, I_{1:N}^{0:t}))+g)/\epsilon))
% \end{array}
% \end{equation}

\begin{equation}
\begin{array}{l}
    \mathcal{G}_i^t = argmax(log(p(\mathcal{G}_i^t|X_{1:M}^{t}))+g) 
\end{array}
\end{equation}
where, $g=-log(-log(u))$, $ u\sim uniform(0,1)$ is the Gumbel noise. The function $argmax(\cdot)$ is not differential, we further use $softmax(\cdot)$ to approximate it. $\mathcal{G}_i^t$  has the shape $(M, G)$ of which each row is an  one-hot vector where element 1 describe the social interaction pattern.  
% , $g$ is a vector of independent and identically distributed samples drawn from $Gumbel(0, 1)$, $\epsilon$ is a hyper-parameter that control the distributions of sampling, here we set $\tau=0.1$. 
% \begin{equation}
% \begin{array}{l}
%     \eta_{i}^{t} = \sum_{j}(\phi_2(\mathcal{G}_j^t;\omega_2^*)*\hat{\mathcal{E}}_{j}^{t} )
% \end{array}
% \end{equation}
% where, $\phi_2$, a fully connected layer with weights $\omega_2^*$, is used to embed interaction types, $\eta_{i}^{t}$ is recurrent over time and  imply the current interaction of agent $i$ with all the other agents.

\subsection{Regulations on Social Patterns}\label{losssocial}
Real-world trajectory-related social behaviors vary on-the-fly and are hard to interpret.  Our hypothesis social patterns can be learned from trajectory prediction directly and then  help  prediction in turn. At the core of our approach is discriminating the social features in a representation space as shown in Figure 4.  Features of similar social behaviors should be clustered together and far away from other social behaviors. Shortly speaking, the features of same clusters contract and features of different clusters contrast.  Inspired by Maximum Coding Rate Reduction ~\cite{chan2022redunet}, we utilize coding rate to describe the compactness of  the features of the clusters. More specifically,  the entire space of features from all clusters should span a space of the largest volume and the coding rate of whole feature set should be as large as possible while the coding rate of the features  of the same clusters should be as small as possible. 

By expanding the size of $\hat{\mathcal{E}}_{i}^{t}$ to $(M, G, R)$ and multiplying it with $\mathcal{G}_i^t$, social features are gathered for each patterns. The process is repeated for each agent pedestrian. After tensor rearrangement,  we obtain $Z^t = \{z_m^t|m=1, 2,\cdots, G\}$ where $z_m$ represent all the social features belong to pattern $m$. The number of features stored in $z_m$ is dynamic during training and testing.   According to ~\cite{ma2007segmentation}, the average coding rate of features of all clusters at any time instance $t$ can be calculated as:
% \begin{equation}
% \begin{array}{l}
% % Requires: \usepackage{amsmath}
\begin{equation}
    R_c(\mathbf{Z}^t, \epsilon \mid \mathbf{\Pi}) = \sum_{m=1}^{G} \frac{\operatorname{tr}(\mathbf{\Pi}^m)}{2M} \log \det \left( \mathbf{I} + \frac{d}{\operatorname{tr}(\mathbf{\Pi}^m) \epsilon^2} \mathbf{Z}^t \mathbf{\Pi}^m {\mathbf{Z}^t}^\top \right)
\end{equation}

Likewise, the coding rate of the entire feature space is calculated as:
\begin{equation}
    R(Z^t, \epsilon) = \frac{1}{2} \log \det \left( \mathbf{I} + \frac{d}{M \epsilon^2} \mathbf{Z}^t {\mathbf{Z}^t}^\top \right)
    \label{eq:equation_placeholder}
\end{equation}
where, $\prod = \{ \prod^{m} \in R^{(M, M)}\}_{m=1}^{G}$ is a set of diagonal matrices whose diagonal entries represent the cluster to which the social interaction between agent and pedestrian $M$ belongs.  $tr(\cdot)$ is the trace of a square matrix, $\epsilon$ is the distortion.  The loss item based on $R_c(\mathbf{Z}^t, \epsilon \mid \mathbf{\Pi})$ and $R(Z^t, \epsilon)$ is:
\begin{equation}
    \mathcal{L}_{social} = - \sum_{t = \tau }^{T-1} \lambda (R(Z^t, \epsilon) - R_c(\mathbf{Z}^t, \epsilon \mid \mathbf{\Pi}))
\end{equation}
$\mathcal{L}_{social}$ allows to learn larger $R(Z^t, \epsilon)$ meaning more distinct cross patterns while smaller $ R_c(\mathbf{Z}^t, \epsilon \mid \mathbf{\Pi})$ meaning a more focused inner pattern, which is preferred in our method.

%     R_c(Z, \epsilon | \prod) = \sum \\
%     q_{in} = \frac{\sum_{m=1}^{G} D(\sum z_m/g_m,z_m) }{G}
% \end{array}
% \end{equation}

% With the similarity measured by cosine distance, an item revealing the how similar the features of the same pattern  $m$ are is proposed:
% \begin{equation}
% \begin{array}{l}
%     q_{in} = \frac{\sum_{m=1}^{G} D(\sum z_m/g_m,z_m) }{G}
% \end{array}
% \end{equation}
% where, function $D(\cdot)$ is cosine similarity, $g_m$ is the number of social features of pattern $m$. The item measuring the similarity of cross-patterns is shown as follows:
% % \begin{equation}
% % \begin{array}{l}
% %     q_{crs} = \frac{\sum_{m=1}^{G} \sum_{n=m+1}^{G}D(\sum z_m/g_m,z_n/g_n) }{C_G^2}
% % \end{array}
% % \end{equation}
% \begin{equation}
% \begin{array}{l}
%     q_{crs} = \frac{\sum_{m=1}^{G} \sum_{n=m+1}^{G}D(\sum z_m/g_m,z_n/g_n) }{0.5*(G(G+1)))}
% \end{array}
% \end{equation}

%   The loss item based on $q_{crs}$ and $q_{in}$ is shown as follows:
% \begin{equation}
% \begin{array}{l}
%     \mathcal{L}_{social} = log\frac{exp(q_{in})}{\alpha \cdot exp(q_{crs})-exp(q_{in})}
% \end{array}
% \end{equation}
% where, $\alpha$ is a hyperparamter. 

\subsection{Pattern Aggregation}\label{aggre}
A simple yet efficient aggregation strategy is introduced by combining social features and  social patterns for path forecasting. Firstly, we encode the social patterns formatted as one-hot vectors. Practically, a fully connected layer is used for $\mathcal{G}_i^t$ of shape $(M, G)$ to get  $\hat{\mathcal{G}}_i^t$ of shape $(M, R)$. Follow the symbol usage of $\hat{\mathcal{E}}_{i}^{t}$, we use $\hat{\mathcal{E}}_{i}^{t}$ to denote social features of agent $i$. By a multiplication then a sum up operation between   $\hat{\mathcal{E}}_{i}^{t}$ and $\hat{\mathcal{G}}_i^t$  over each corresponding pattern, an aggregated feature $\eta_i^t$ specified for agent $i$ is obtained.
\begin{figure*}
    \centering
    \includegraphics[clip=true, viewport=50 180 700 420, width=0.75\textwidth]{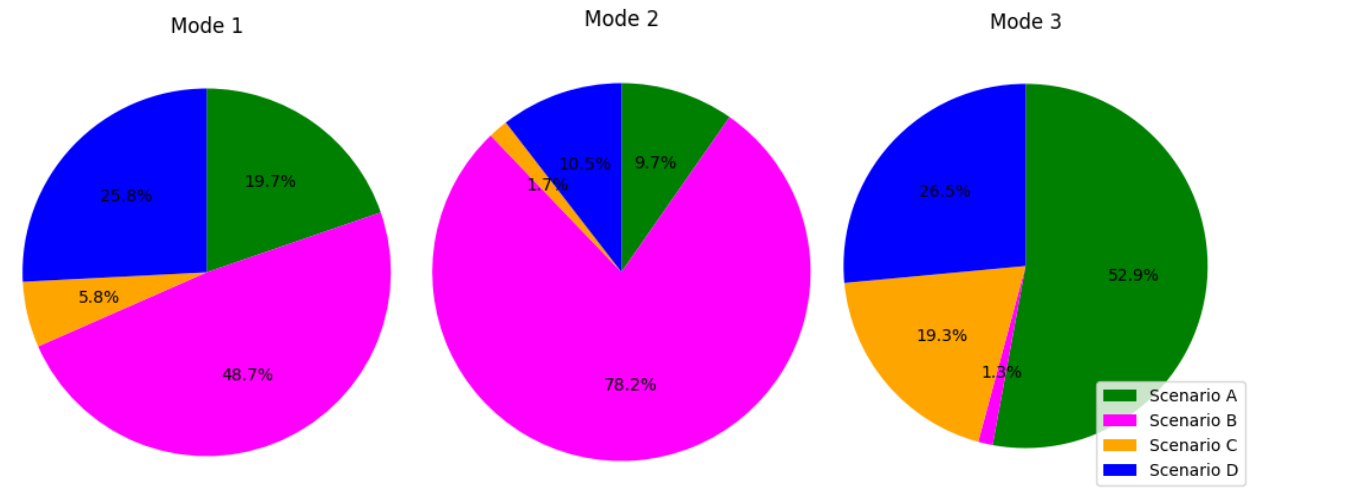}
    \caption{Interpretation on Mode. We explain modes by analyzing  them from several typical interaction scenarios that are designed based on common sense. }
    \label{fig:mode}
\end{figure*}
\subsection{Path Forecasting}\label{path}
As mentioned before, the agent $i$ at time instance $t$ is characterized with location $(x_i^t, y_i^t)$ and offset $(u_i^t, v_i^t)$. Following most of the current research,  we use offset for trajectory prediction, which not only stabilize the whole training process but also improve the prediction performance. A fully connected layer with ReLU non-linearity is applied to $(u_i^t, v_i^t)$ and get features $f_i^{t}$. $f_i^{t}$ and $\eta_i^t$ is added then for predicting the offset state of next time step.
% \begin{equation}
% \begin{array}{l}
%     f_i^{t} = \phi_3((u_i^{t}, v_i^{t});w_3^*)
% \end{array}
% \end{equation}
% where $\phi_2(\cdot)$ and $\phi_3(\cdot)$ are fully connected layers with ReLU non-linearity, $w_2^*$ and $w_3^*$ are the embedding weights. We represent social state for agent $i$ at time $t$ as $s_i^{t}$ and concatenate it with $f_i^{t-1}$ to predict next state of agent.
\begin{equation}
\begin{array}{l}
    h_i^{t} = \phi(h_i^{t-1}, f_i^{t} + \eta_i^{t};w_h^*)
\end{array}
\end{equation}
where, $\phi(\cdot)$ is the trajectory encoder. Here, we use LSTM and its weights $w_h^*$ are shared between all people in a scenario. Trajectory decoder also share the same weight with encoder while encoder is for encoding trajectory observations and decoder is for futural prediction.  To capture the multi-modality of future paths, we utilize mixture density network(MDN) that combines a multilayer perception with GMMs. The next location of agent conditioned on hidden states of LSTM $h_i^{t}$ are denoted as:

\begin{equation}
\begin{array}{l}
    p(\hat{Y}_i^{t+1}|h_i^{t}) = \sum_{g=1}^{B}\alpha_g^{t} p(\hat{Y}_i^{t+1}|\mu_g^{t}, \sigma_g^{t})
\end{array}
\end{equation}
where $B$ is the number of Gaussian models of MDN, $\alpha_g^{t}$  is the prior of $gth$ kernel, $p(\hat{Y}_i^{t+1}|\mu_g^{t}, \sigma_g^{t})$ is the probability density functions (PDFs) given by $gth$ component of GMMs which is a bivariate Gaussian model parametrized by the mean $\mu_g^{t} = (\mu_x, \mu_y)_g^{t}$, standard deviation $\sigma_g^{t} = (\sigma_x, \sigma_y)_g^{t}$ and correlation coefficient $\rho_g^{t}$.  We set $\rho_g^{t}$ as constant and learn $\mu_g^{t}$, $\sigma_g^{t}$ and $\alpha_g^{t}$through our network.
\begin{equation}
\begin{array}{l}
    \alpha_g^{t} = \frac{exp(\textsl{a}_g^{t})}{\sum _{k=1}^M exp(\textsl{a}_k^{t})} \\
    \mu_g^{t} = \textsl{u}_g^{t} \\
    \sigma_g^{t} = exp(\textsl{z}_g^{t})
\end{array}
\end{equation}
where $\{\textsl{a}_g^{t}|g = 1,\cdots, B\}$, $\{\textsl{u}_g^{t}|g = 1,\cdots, B\}$ and $\{\textsl{z}_g^{t}|g = 1,\cdots, B\}$is obtained by applying fully connected layers $\phi_\alpha(\cdot)$, $\phi_\mu (\cdot)$ and $\phi_\sigma(\cdot)$ to $h_i^t$ respectively.
\subsection{Loss Function}\label{loss}
The loss function is  designed to compute 
negative log-likelihood of future trajectories over all components of a mixture model (Eq. (10)) plus social loss. 
\begin{equation}
\begin{array}{l}
    \mathcal{L}_{mdn} = - \sum_{t = \tau }^{T-1} log(\sum_{g=1}^{M}   \alpha_g^{t} p(\hat{Y}^{t+1}| \mu_g^{t}, \sigma_g^{t})   )\\
        \mathcal{L}_{all} = \mathcal{L}_{mdn} + \beta \cdot \mathcal{L}_{social}
\end{array}
\end{equation}

% To capture the variety of multi-modes and truly learn the multi-modality of human motion, we design a WTA loss as Eq. (9). In the training process, we always base the winner selection on the probability. We compute loss by multiply the winner probability with learned weight. The weight of winner mode increase through training. 

% \begin{equation}
% \begin{array}{l}
%     \mathcal{L}_{wta} = - \sum_{t = \tau}^{T-1} log(\alpha_g^{t} p(\hat{Y}^{t+1}| \mu_g^{t}, \sigma_g^{t}))\\
%     g = \underset{g}{\arg \max} \, \, p(\hat{Y}^{t+1}| \mu_g^{t}, \sigma_g^{t})\\
% \end{array}
% \end{equation}
% Finally, social loss and $\mathcal{L}_{wta}$ are added for training the proposed model end to end. 

% \begin{equation}
% \begin{array}{l}
%     \mathcal{L}_{all} = \mathcal{L}_{mdn} + \beta \cdot \mathcal{L}_{social}
% \end{array}
% \end{equation}

\section{Experiments}
In this section, we try to answer 3 questions: (1) what do modes mean, (2) when do modes shift, (3) will the social interaction clustering improve the prediction accuracy. Moreover, the proposed model is evaluated on two publicly available datasets: UCY~\cite{lerner2007crowds} and ETH~\cite{pellegrini2009you}. The two datasets contain 5 sets, which are UCY-zara01, UCY-zara02, UCY-univ, ETH-hotel, ETH-eth in 4 crowded scenarios with totally 1536 trajectories. We firstly preprocess those two datasets by resampling them as 2.5fps and transforming the coordinates of people to world coordinates in meters. 

\textbf{Implementation Details}.
The experiments are implemented using Pytorch under Ubuntu 16.04 LTS with a GTX 4090 GPU. The size of hidden states of LSTM is set to 128.  The embedding layers are composed of a fully connected layer with size  128. The batch size is set to 8 and all the methods are trained for 200 epochs. 
The optimizer RMSprop is used to train the proposed model with learning rate 0.001. We clip the gradients of LSTM with a maximum threshold of 10 to stabilize the training process. The model outputs GMMs with five components.  $\beta$ is set as 0.1. Specifically, we set $G$ as 3 which is also compliant with the other research ~\cite{marewski2010good,girase2021loki,moussaid2011simple} which state that while in the ideal case of
having infinite rationality, humans are expected to make better decisions through more complex cognition, the limited
time, knowledge and computational power of humans enhances the desirability of simple cognition for more robust
performance to cope with the uncertainty of the world. In the context of crowd walking, where rapid decision-making is required, we argue that social interactions tend to fall into only a few  categories.

\textbf{Evaluation Approach}.
The proposed model is trained and tested on the two datasets with leave-one-out approach: trained on four sets and tested on the remaining set. We observe the trajectories
for 8 timesteps (3.2 sec) and show prediction
results for 12 timesteps (4.8 sec). To evaluate the performance, we compare our method with other state-of-the-art models on two generally used metrics. 

1. Average displacement error (ADE): average L2 distance over all  prediction results and ground truth. ADE measures average error of the predicted trajectory sequence.

2. Final displacement error (FDE): distance between prediction result and ground truth at final timestep. FDE measures the error "destination" of the prediction.

% \begin{equation}
% \begin{aligned}
%    log p(Y|X) = log \sum_{Z} p(Y, Z|X) = \sum_{Z}p(Z|X, Y) log\frac{p(Y, Z|X)}{p(Z|X, Y)} 
% \end{aligned}
% \end{equation}
% Optimizing for the equation directly is difficult as the posterior distribution $p(Z|X, Y)$ (i.e. the true distribution of social interaction patterns) is high-dimensional, complicated, and hard to compute. To address it, we  use $q(Z|X, Y)$ to approximate it and then use Jensen's inequality to make the equation arrive at the evidence lower
% bound (ELBO) as follows:
% % \begin{equation}
% % \begin{aligned}
% %    log p(Y|X)  &= \sum_{Z}q(Z|X, Y) log\frac{p(Y, Z|X)}{q(Z|X, Y)}  + KL(q||p) \\
% %    &\ge \sum_{Z}q(Z|X, Y) log p(Y, Z|X) + KL(q||p)
% % \end{aligned}
% % \end{equation}
%  We learn the model parameters by maximizing the variational lower bound on the data log-likelihood. When the model converges,  the approximation distribution  $q(Z|X, Y)$ is considered to represent the distribution of true distribution $p(Z|X, Y)$.  

%  Moreover, the proposition of our inference of the social interaction patterns is to cluster/classify the social interactions into groups for better interpretability, which is achieved by  $q(Z|X, Y)$ further learned from data.
\begin{figure*}
    \centering
    \includegraphics[clip=true, viewport=20 80 950 500, width=0.85\textwidth]{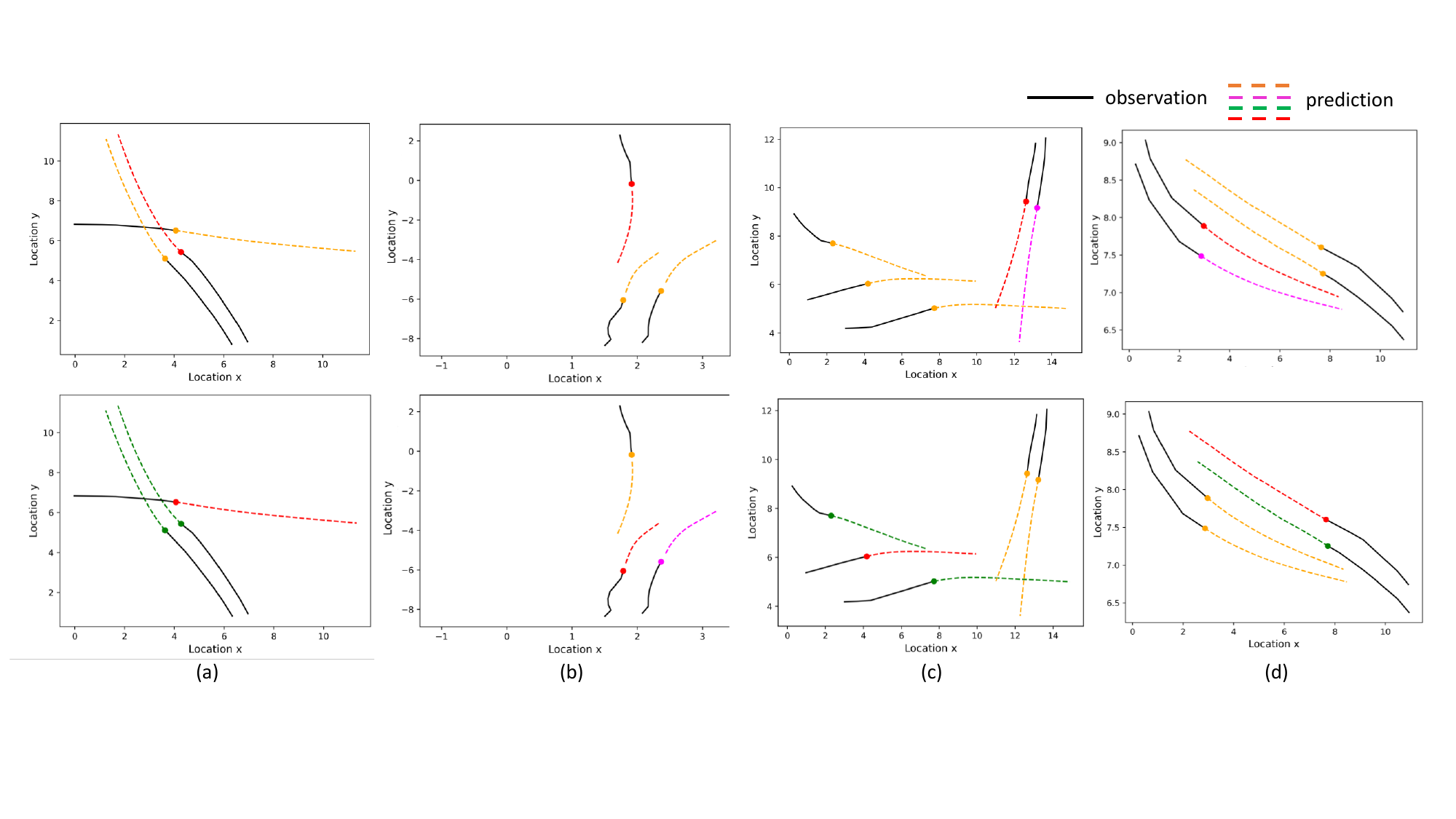}
    \label{fig:mode}
    \caption{Visualization on modes. Mode 1-green, Mode 2-Magenta, Mode 3-Orange. The red dot and red lines represent the agent pedestrian while the other neighbor pedestrians' color show their social interaction mode with the agent which is how agent socially interacted with his/her neighbors.   }
\end{figure*}
\subsection{What do modes mean?}
Since the clustering of social interactions is performed without labels, interpreting the meaning of each mode is inherently challenging. To provide an intuitive explanation of the clustered categories, we define several typical interaction scenarios based on common sense. \textbf{Scenario A} describes two pedestrians approaching each other from opposite directions at a close distance, a situation that is more likely to involve intense and aggressive interaction. \textbf{Scenario B} corresponds to two pedestrians walking in parallel in the same direction at a close distance, which typically reflects following behavior and can be regarded as mild interaction or even non-interaction. \textbf{Scenario C} represents two pedestrians approaching from opposite directions but starting at a relatively large distance, which tends to result in all kinds of social interactions which are mild interaction, no interaction or aggressive social interaction. \textbf{Scenario D} describes two pedestrians walking back-to-back in opposite directions with a large separation, a situation that usually indicates no interaction. We performed a statistical analysis of social interactions in the dataset. The results are presented in the Figure 5. It can be observed that \textbf{Mode 3} is predominantly distributed in Scenario A, indicating that social interactions belonging to this mode are more likely to involve \textbf{aggressive interactions}. In contrast, \textbf{Mode 2} is more concentrated in Scenario B, suggesting that it tends to exhibit \textbf{mild forms of interaction}. It is worth noting that \textbf{Mode 1} is relatively evenly distributed across all four scenarios, implying that this mode represents a more common form of interaction that is less influenced by pedestrians’ relative positions or walking directions. Therefore, we recommend interpreting mode 1 as corresponding to \textbf{non-interaction} or very slight social interactions.

To further gain insight into the meaning of each mode, i.e., the interpretation of the clusters, we conducted a visualization across all datasets, with several representative examples shown in the Figure 6. We observe that social interactions within a group are not always symmetric. For instance, in Figure (a1) (the upper subfigure of panel a), the agent exhibits aggressive interactions with its two neighbors, while in Figure (a2) the agent is already moving away from its neighbors and therefore pays less attention to them. In Figure (b2), the agent primarily attends to the pedestrian on its left, but at the same time interacts with the neighbor on the right as it intends to make a right turn. The scenarios in Figures (c) and (d) involve a larger number of pedestrians, yet the interaction behaviors remain consistent with those observed in other cases. This further demonstrates the rationality of the learned modes by our proposed method.
  
\subsection{When do modes shift?}
\begin{figure*}
    \centering
    \includegraphics[clip=true, viewport=10 275 930 480, width=1.0\textwidth]{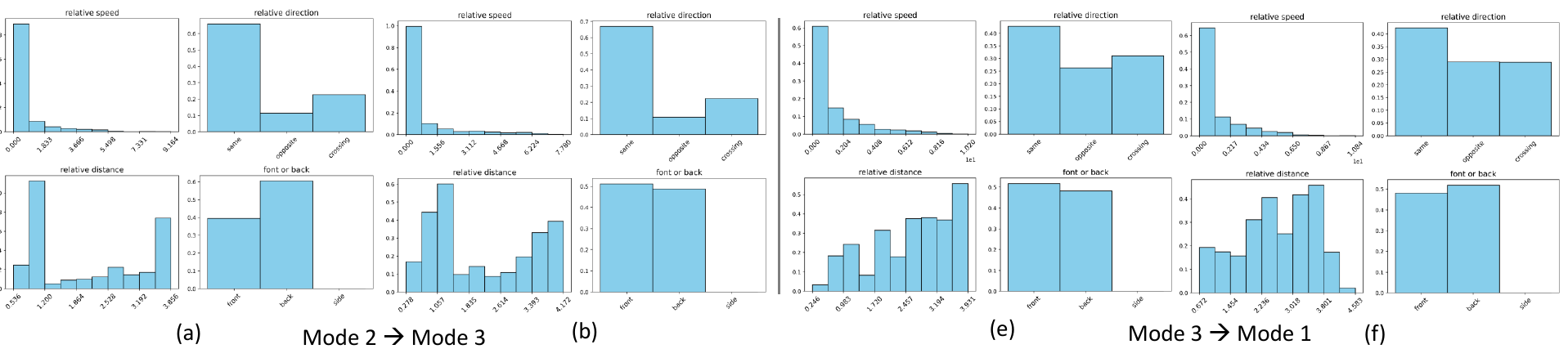}
    \label{fig:mode}
    \caption{Statistics on mode change. We analyze the change on  relative speed(upper left), walk direction(upper right), distance(lower left) and position relation(lower right) respectively when mode change from 2 to 3 and from 3 to 1.  }
\end{figure*}
% \subsection{How mode affect the trajectory prediction?}
We further analyze the short time periods when a mode transition occurs, aiming to understand which relational factors between the agent pedestrian and its neighbors change, thereby providing insights into the elements that drive the transition. Due to space limitations, we only present the cases of Mode 2 → Mode 3 and Mode 3 → Mode 1 in Figure 7. We design four indicators for this analysis: (1) relative speed magnitude, (2) walking direction (i.e., same, opposite, crossing), (3) relative distance, and (4) the relative position of neighbors with respect to the agent (i.e., front, back, side).

Our analysis reveals that when the social interaction becomes more intense (e.g., Mode 2 → Mode 3), the most significant change lies in the relative positional relationship, as more neighbors appear in front of the agent. At the same time, the relative speed magnitude increases and the relative distance decreases, while the relative walking direction does not change substantially. In contrast, when the social interaction gradually becomes milder (e.g., e → f in Figure 7), more neighbors move to the rear of the agent, and the walking direction shifts more often from opposite to same or crossing. Correspondingly, the relative distance slightly increases, while the relative speed magnitude slightly decreases. Based on the above analysis, we may argue that the relative positional relationships between pedestrians are one of the main factors driving mode transitions.

\subsection{Mode Accuracy}
To further illustrate the significance of learning clustered social interactions, we also evaluate their impact on trajectory prediction accuracy. Table ~\ref{results1} presents a comparison of the proposed method against baseline approaches in terms of prediction accuracy. Linear methods underperform because they cannot capture social context or the multi-modality of human motion. Social LSTM performs on par with a vanilla LSTM, while offset-based LSTM inputs stabilize training and improve accuracy. Methods such as Sophie, Social GAN, and Social BiGAT, which model long-term uncertainty, achieve superior results. “20 samples” denotes selecting the best outcome from 20 samples drawn from the predicted distribution.  Interpretability inevitably reduces the accuracy of prediction models. However, surprisingly overall, our model outperforms the baselines by a large margin on the ETH datasets, demonstrating its strong capability to capture complex interaction dynamics in crowded and unconstrained environments. On the UCY datasets, our method achieves performance comparable to other state-of-the-art baselines. These results highlight the robustness and generalizability of the proposed model across different benchmark datasets.

\begin{table}[!t]
\renewcommand{\arraystretch}{1.3}
\centering
\caption{Quantitative results of baselines vs. our method across datasets for predicting 12 future timesteps(4.8 sec) given 8 timesteps observation(3.2 sec).}
\label{results1}
\begin{adjustbox}{width=0.5\textwidth}
\small
\begin{tabular}{ |c||  c| c| c| c| c| c| c|}
\hline
\multicolumn{1}{|c||}{\multirow{2}{*}{Method} } &
\multicolumn{1}{|c|}{\multirow{2}{*}{Note}} &
\multicolumn{6}{|c|}{Evaluation\,\,(ADE(m)/FDE(m))} \\ \cline{3-8}
\multicolumn{1}{|c||}{} &\multicolumn{1}{|c|}{} &ETH-eth &ETH-hotel &UCY-univ &UCY-zara01 &UCY-zara02 & AVG \\
\hline
Linear & kalman filter & 1.65/2.84 & 0.99/1.70 & 0.86/1.51 & 0.83/1.44 & 0.54/0.96 & 0.97/1.69 \\
\hline
LSTM & offset as input & 0.71/1.40 & 1.15/2.09 & 0.72/1.49 & 0.48/0.98 & 0.38/0.77 & 0.69/1.35 \\
\hline
Social LSTM \cite{alahi2016social} & social pooling & 1.09/2.35 & 0.79/1.76 & 0.67/1.40 & 0.47/1.00 & 0.56/1.17 & 0.72/1.54 \\
\hline
Sophie\cite{sadeghian2019sophie} & 20 samples & 0.70/1.43 & 0.76/1.67 & 0.54/1.24 & 0.30/0.63 & 0.38/0.78 & 0.54/1.15 \\
\hline
Social GAN\cite{gupta2018social} & 20 samples & 0.72/1.29 & 0.48/1.01 & 0.56/1.18 & 0.34/0.69 & 0.31/0.65 & 0.48/0.96 \\
\hline
Social BiGAT\cite{kosaraju2019social} & 20 samples & 0.69/1.29 & 0.49/1.01 & 0.55/1.32 & 0.30/0.62 & 0.36/0.75 & 0.48/1.00 \\
\hline
Social STGCNN\cite{mohamed2020social} & 20 samples & 0.64/1.11 & 0.49/0.85 &\textbf{0.44/0.79} & 0.34/\textbf{0.53} & 0.30/0.48 & 0.44/0.75
\\
\hhline{|=||=|=|=|=|=|=|=|}
Model-V1 & 20 samples & \textbf{0.33/0.68} & \textbf{0.16/0.36} & 0.60/1.22 & \textbf{0.30}/0.66 & \textbf{0.27/0.45} & \textbf{0.33}/\textbf{0.67}\\
\hline
\end{tabular}
\end{adjustbox}
\end{table}

\subsection{Ablation study }
To explain how our model works, we also represent results (Table 2) of various versions of our models in an ablative setting by 
Model-V1 and Model-V2: the whole version of our model with different sampling times, 
Model-V3: with the entire model, mdn loss without social loss, i.e. without regulations on modes, 
% with sample times of 20 and 1 respectively,
Model-V4: with the entire loss and uses attention mechanism to generate social states without mode extraction, 
Model-V5: with Mean Squared Error (MSE) loss, with mode extraction and applies softmax over each mode before integration. 

By comparing Model-V2 and Model-V3, we observe that introducing constraints into the clustering process is beneficial, as it significantly improves prediction accuracy. Moreover, such constraints help ensure that clustering in the high-dimensional feature space is more reliable. The comparison between Model-V3 and Model-V4 further demonstrates the effectiveness of extracting interaction patterns and integrating them into the prediction task. While most existing approaches have not quantitatively investigated social interactions, our study makes an initial attempt in this direction. Similarly, the comparison between Model-V3 and Model-V5 confirms the effectiveness of the MDN loss.

\begin{table}[!t]
\renewcommand{\arraystretch}{1.3}
\centering
\caption{Ablation Study on Our Model. M indicate if we learn mode for social interactions or not, K means sampling times, Loss present if we use social loss or not.}
\label{results2}
\begin{threeparttable}
\begin{adjustbox}{width=0.5\textwidth}
\small
\begin{tabular}{ |c||  c| c| c|| c| c| c| c| c| c|}
\hline
\multicolumn{1}{|c||}{\multirow{2}{*}{Variant ID} } &
\multicolumn{3}{|c||}{Components} &
\multicolumn{6}{|c|}{Evaluation\,\,(ADE(m)/FDE(m))} \\
\cline{2-10}
&M & K & Loss 
&ETH-eth &ETH-hotel &UCY-univ &UCY-zara01 &UCY-zara02 & AVG \\

\hhline{|=||=|=|=||=|=|=|=|=|=|}
Model-V1 &$\surd$ &20 &$\mathcal{L}_{all}$ & \textbf{0.33/0.68} & \textbf{0.16/0.36} & \textbf{0.60/1.22} & \textbf{0.30}/\textbf{0.66} & \textbf{0.27/0.45} & \textbf{0.33}/\textbf{0.67}\\
\hline
Model-V2 &$\surd$ &1 &$\mathcal{L}_{all}$ & 0.46/1.12 &0.36/1.00 &0.70/1.51 & 0.44/1.09 & 0.32/0.76& 0.46/1.10 \\
\hline
Model-V3 & $\surd$ &1 &$\mathcal{L}_{mdn}$ & 0.47/0.90&  0.26/0.46 & 0.68/1.34 & 0.37/0.68 & 0.29/0.51&  0.41/0.78
 \\
\hline
Model-V4 &  &1 &$\mathcal{L}_{mdn}$ & 0.77/1.68& 0.47/0.92 &0.74/1.51 &0.58/1.23& 0.36/0.76 &0.58/1.22 \\
\hline
% Model-V5 &$\surd$ &1 &$\mathcal{L}_{mdn}$ &0.65/1.33 & 0.49/1.01 & 0.75/1.55 & 0.61/1.33 & 0.37/0.78 & 0.57/1.20 \\
% \hline
Model-V5 &$\surd$ &1 &$\mathcal{L}_{2}$ & 0.60/1.21 & 0.39/0.95 & 0.69/1.50 & 0.47/1.14 & 0.40/0.80 & 0.51/1.12\\
\hline
\end{tabular}
\end{adjustbox}
\end{threeparttable}
\end{table}

\section{Conclusion}
In this work, we proposed a latent-variable generative approach to interpret social interactions in pedestrian walking. The latent variable is learned from high-dimensional features of social interactions, while a maximum coding rate reduction criterion is employed to constrain the feature space and construct a social loss function. The clustering process is not hand-crafted but totally data-driven. Experimental results demonstrate the effectiveness of our method in capturing and explaining interaction patterns among pedestrians. For future work, we plan to apply our approach to a wider range of datasets and conduct more extensive experiments to further evaluate its generalization ability. We are also going to explore the number of clusters, i.e. the social interaction modes. Besides, interpretability inevitably reduces the accuracy of prediction models. we will explore more effective ways of incorporating interaction patterns into a broader range of tasks.


% \bibliographystyle{cas-model2-names}

\bibliographystyle{plain}

\bibliography{acmart}

% % Biography
% % Here goes the biography details.

% % Here goes the biography details.

\end{document}